\documentclass[11pt]{article}
\usepackage{coling2020}
\usepackage{times}
\usepackage{url}
\usepackage{latexsym}
\usepackage{multirow}
\usepackage{xcolor}

\usepackage{setspace}
\usepackage{caption}

\colingfinalcopy % Uncomment this line for the final submission

\title{Towards Preemptive Detection\\of Depression and Anxiety in Twitter}

\author{David Owen \hspace{0.8cm} Jose Camacho-Collados \hspace{0.8cm} Luis Espinosa-Anke \\
         School of Computer Science and Informatics \\
	    Cardiff University, United Kingdom\\
	     {\tt \{owendw1,camachocolladosj,espinosa-ankel\}@cardiff.ac.uk}}

\date{June 2020}

\begin{document}
\maketitle

\begin{spacing}{0.98}

\blfootnote{This work is licensed under a Creative Commons Attribution 4.0 International Licence. Licence details:\newline \url{http://creativecommons.org/licenses/by/4.0/.}}

\begin{abstract}
%\begin{center}
Depression and anxiety are psychiatric disorders that are observed in many areas of everyday life. For example, these disorders manifest themselves somewhat frequently in texts written by non-diagnosed users in social media. However, detecting users with these conditions is not a straightforward task as they may not explicitly talk about their mental state, and if they do, contextual cues such as immediacy must be taken into account. When available, linguistic flags pointing to probable anxiety or depression could be used by medical experts to write better guidelines and treatments. In this paper, we develop a dataset designed to foster research in depression and anxiety detection in Twitter, framing the detection task as a binary tweet classification problem. We then apply state-of-the-art classification models to this dataset, providing a competitive set of baselines alongside qualitative error analysis. Our results show that language models perform reasonably well, and better than more traditional baselines. Nonetheless, there is clear room for improvement, particularly with unbalanced training sets and in cases where seemingly obvious linguistic cues (keywords) are used counter-intuitively.
%\end{center}
\end{abstract}

\section{Introduction}
Mental illnesses are psychiatric disorders that may cause sufferers significant distress and impair their ability to function in social and work activities \cite{BoltonDerek2008Wimd}. The most prevalent mental illnesses are depression and anxiety, which are estimated to affect nearly one in ten people worldwide (676 million) according to a recent study \cite{world2016world}. While depression and anxiety are different disorders, they also share symptoms and, thus, clinicians often diagnose patients with both illnesses at consultation.\footnote{https://www.bupa.co.uk/newsroom/ourviews/2017/10/anxiety-depression} Identifying these conditions at early stages is relevant not only because of their inherent importance, but also because they are precursors to major related concerns in public health including self-harm \cite{centers2015suicide}, making timely diagnosis and treatment even more essential. However, sufferers of depression and anxiety can find that it takes great courage and strength to seek professional treatment \cite{miller2016}. They may also be afraid to confide in their peers due to the stigma of mental illness \cite{wasserman2012suicide}. 

With reluctance to seek professional treatment or rely on their peers, sufferers often turn to online resources for support. These include both specialised and general communities, with Twitter and Reddit \cite{yates2017depression} being paramount examples of the latter. Because of this, systems that can automatically detect and flag such cases at a large-scale are highly desirable \cite{guntuku2017detecting}. They may enable prompt analysis and treatment, which is crucial in the early development of such conditions. Moreover, the interpersonal and economic effects of these illnesses may be mitigated with prompt intervention \cite{lexis2011prevention}.

%(see Section \ref{annotation} for details about the annotation process)
In this paper, we build a classification dataset\footnote{The datasets and code used in our experiments are available online at the following repository:\\
\url{https://bitbucket.org/nlpcardiff/preemptive-depression-anxiety-twitter}.
} to assist in the detection of depression and anxiety in Twitter, and compare several text classification baselines. The results show that state-of-the-art language models (LMs henceforth) like BERT \cite{BERT} unsurprisingly outperform competing baselines. However, when the dataset shows an unbalanced distribution, linear models perform on par.  Finally, alongside quantitative results, we also provide a qualitative analysis through which we aim to better understand the strengths and limitations of the models under study. Further, we identify the linguistic patterns alluding to the presence of depression and anxiety that elude all of the classifiers, and consider how we might improve performance against such patterns in the future.

\section{Related Work}
\label{related}

\newcite{baclic2020challenges} surveyed the use of NLP in healthcare and identified the inherent opportunities and challenges. NLP permits speedy analysis of large volumes of unstructured text such as electronic patient records or social media posts, which can help support early healthcare interventions. For example, automated analysis of consumers' online reviews was used to predict the presence of depression in reviewers prior to its formal diagnosis \cite{harris2014health}. %\red{TO REPHRASE AND CHECK TATMAN REFERENCE: 
However, Baclic et al. noted that the effectiveness of NLP in the domain of mental health is constrained by a lack of high quality training data. Consequently, we chose to build our own labelled dataset.

In terms of detecting depression and anxiety in social media, the work of \newcite{yates2017depression} is perhaps the most related to ours. They showed that depression among Reddit participants can be detected by identifying certain lexical and psycholinguistic features in the contents of their Reddit postings. These features include indications of negative psychological processes such as anger and sadness, which may be denoted by the words "hate" and "grief" respectively \cite{Pennebaker2015TheDA}. The work of Yates et al. is therefore highly relevant to ours since it concerns automated detection of mental illness in written social media discourse. However, there are three key aspects that set our paper apart, namely: (1) 
%\begin{enumerate}
    %\item 
We identify users of social media platforms who appear to have \emph{not} yet been diagnosed with mental illness. We do this in the spirit of providing the opportunity for early healthcare intervention in these cases; 
    %\item 
(2) we consider automated classification using feature rich representations of users' online postings using state-of-the-art NLP methods such as word embeddings and pre-trained LMs; and % \cite{wolf2018sentenceembeddings}.
    %\item 
(3) we consider concise written discourse in the form of tweets, rather than Reddit postings, which are generally more verbose.
%\end{enumerate}

\section{Dataset Construction}
\label{dataset-construction}
In this section, we describe our process for building a dataset for detecting depression and anxiety in Twitter. We describe the tweet collection procedure (Section 3.1), annotation (Section 3.2), and provide information about the inter-annotator agreement (Section 3.3).

\subsection{Tweet collection}
First, we used Twitter's Stream API to compile a large corpus of tweets. %Twitter corpus of 60M tweets. 
%It comprised 584 million tokens amounting to 3.6 gigabytes of uncompressed text. 
All tweets were of English language and published between May 2018 and August 2019.\footnote{Twitter's automatic language labelling was used to identify English tweets.} We only considered tweets containing at least three tokens and without URLs so as to avoid bot tweets and spam advertising. All personal information, including usernames (denoting the author or other users) and location were removed from the corpus - only textual information was retained. We did however retain emojis and emoticons. We surmised that they may, in part at least, be indicative of depression or anxiety.

The corpus was then filtered. We aimed to identify tweets whose authors may be suffering from depression or anxiety but may not yet have been diagnosed by a clinician. To achieve this, we sought tweets containing occurrences of \emph{depress}, \emph{anxie}, or \emph{anxio}, but not \emph{diagnos}\footnote{e.g. "My anxiety is terrible today"} - an approach similar to that used by \newcite{bathina2020depressed}. This produced an initial set of 89,192 tweets. From these tweets we proceeded to annotate a random subset of 1,050 tweets to arrive at our dataset.

\subsection{Annotation}
\label{annotation}

Three human annotators were appointed. The prerequisites for these annotators were to be fluent in English and to have familiarity with Twitter. The 1,050 tweet dataset was divided into three distinct subsets of 300 and one distinct shared subset of 150. Each annotator received one of the former subsets in addition to the latter subset. They were tasked with labelling the 450 tweets that they had received.

One of two numerals was selected by the annotator with respect to each tweet:
\begin{itemize}
    \item[] \textbf{1}: The tweeter appears to be suffering from depression or anxiety.
    \item[] \textbf{0}: The tweeter does not appear to be suffering from depression or anxiety.
\end{itemize}

Guidelines were compiled to aid the annotation exercise. Their purpose was to ensure a consistent approach amongst annotators and to resolve ambiguous cases. These guidelines are defined below along with examples and their suggested labels:
\begin{enumerate}
    \item The tweeter states that they have depression or anxiety\\
    Example: \emph{``I feel sick to my stomach, I hate having such bad anxiety''} - \textbf{1}
    \item The tweeter states that they have had depression or anxiety in the past\\
    Example: \emph{``Counselling fixed my depression''} - \textbf{0}
    \item The tweeter is referring to a fellow tweeter who may have depression or anxiety\\
    Example: \emph{``@user I wish you all the best in beating your anxiety''} - \textbf{0}
    \item The tweeter is temporarily depressed or anxious due to a short or superfluous event\\
    Example: \emph{``Nothing gives me anxiety more than the tills at Aldi''} - \textbf{0}
    \item The tweet is ambiguous or does not provide definitive information\\
    Example: \emph{``Depression is not taken serious enough''} - \textbf{1} %\red{Late comment to discuss, it may make sense to label these instances as zero indeed (even if before I argued for the opposite, my bad!)}
\end{enumerate}

Guideline 5 recommends positive labelling in ambiguous cases. This is to help achieve high recall in terms of tweeters who appear to be suffering from depression or anxiety. Whilst this approach will inevitably retrieve negative instances, an eventual real-world application would require all retrieved instances to be verified manually by medical experts, and therefore high recall at the expense of lower precision is an acceptable tradeoff. In fact, it is recommended that results from automatic classifiers used in healthcare settings should be verified via an ``expert-in-the-loop approach'' \cite{holzinger2016interactive}.

The guidelines evolved following the annotators' first attempts at the exercise. A conflict resolution meeting revealed that while agreement was due to be acceptable, there were instances where annotators felt unable to assign either label. This gave rise to the addition of guideline 5, which allowed the annotators to complete the exercise with confidence.

\subsection{Inter-Annotator Agreement}

Once the annotation was completed, we calculated the Average Pairwise Percentage Agreement of the three annotators with respect to the 150 common tweets that they had received (Table 1).

\begin{table}[t!]
\begin{tabular}{ | c | c | c ||  c | c | c | } 
\hline
\shortstack{Annotators\\1 and 2} &
\shortstack{Annotators\\1 and 3} &
\shortstack{Annotators\\2 and 3} &
\shortstack{Average Pairwise\\Agreement} &
\shortstack{Fleiss' Kappa} &
\shortstack{Krippendorff's Alpha}\\
\hline
78.67 & 80.67 & 80.67 & 80.00 & 0.60 & 0.60\\ 
\hline
\end{tabular}
\caption{Pairwise Percentage Agreements and Inter-Annotator reliability.}
\label{table:1}
\end{table}

An average pairwise agreement of 80\% was recorded. To validate the quality of the exercise two further measures of inter-annotator reliability were selected: Fleiss' Kappa and Krippendorf's Alpha. They are apt for inter-annotator exercises involving more than two annotators \cite{zapf2016measuring}. Both measures returned scores indicating ``substantial agreement'' amongst the annotators \cite{xie2017reformulation}.

Confidence in the annotation guidelines was therefore established versus the 150 tweets common to each annotator. Disagreements in the labels were decided by majority voting among the three annotators. For example, the tweet \emph{``My seasonal depression automatically begun tonight at 12am''} was labelled \textbf{1} by two of the annotators and labelled \textbf{0} by the third annotator. However, majority voting meant that it was finally labelled \textbf{1}. The annotators then proceeded to label their distinct 300 tweet subsets independently.

\section{Experimental evaluation}
In the following we detail the experimental setting (Section \ref{experimental-setting}) and then present the results (Section \ref{results}) and an analysis (Section \ref{analysis}).

\subsection{Experimental setting}
\label{experimental-setting}

\subsubsection{Data}
\label{data}

We prepared our annotated dataset described in Section \ref{dataset-construction} for input to a series of supervised classifiers. The three distinct subsets of 300 annotated tweets were combined to form a training set of 900 tweets. The 150 tweets labelled by all annotators formed the test set. We named this dataset \textit{DATD} (Depression and Anxiety in Twitter Dataset). The test set's ratio of positive instances to negative instances was exactly 1:1 following the annotation exercise. This contrasts with related published datasets upon which no annotation had been performed and all instances were deemed mental illness-related.\footnote{https://github.com/AshwanthRamji/Depression-Sentiment-Analysis-with-Twitter-Data/blob/master/tweetdata.txt}

Following a similar approach to \newcite{bathina2020depressed}, we also compiled a non-annotated set of 3,600 random tweets which did not contain any occurrence of \emph{depress}, \emph{anxie}, \emph{anxio}, or \emph{diagnos}. These were merged with the 900 tweet training set to form a larger training set of 4,500 tweets. The purpose of this large training set (DATD+Rand henceforth) was to recreate a more realistic (and noisy) setting where most training instances are negative. This meant that only 10.5\% of the instances in this training set contained any of the keywords used to compile the positive examples. The 150 tweets labelled by all annotators formed the test set once again. The main characteristics of the two datasets are summarised in Table 2.
\begin{table}[t!]
\centering
\begin{tabular}{|l|cc|c|}
\cline{2-4}
 \multicolumn{1}{c|}{} & \multicolumn{2}{c}{\textbf{Training}} & \multicolumn{1}{|c|}{\textbf{Test}}\\
\cline{2-4}
\multicolumn{1}{c|}{}  & \multicolumn{1}{l}{DATD} & \multicolumn{1}{l}{DATD+Rand} & \multicolumn{1}{|l|}{DATD}\\
\hline
Positive Instances & 473 & 473 & 75\\
Negative Instances & 427 & 4,027 & 75\\
\hline
\hline
Total Instances & 900 & 4,500 & 150 \\
\hline
\end{tabular}
\caption{Characteristics of the datasets used in the evaluation.}
\label{table:2}
\end{table}

\subsubsection{Comparison systems}

We evaluated several binary classifiers on both the DATD and DATD+Rand datasets, guided by existing research concerning problems similar to the one at hand. % The reason for using more than one classifier is due to the ``no free lunch theorem'', which states that there is no one classifier that is ideal for every problem \cite{wolpert1996lack}. The selection of classifiers was guided by existing research concerning problems similar to the one at hand. 
To this end, we deemed a Support Vector Machine (SVM) and an LM to be suitable classifiers. SVMs have demonstrated effectiveness when used with Twitter datasets in healthcare contexts \cite{prieto2014twitter,han2020application}. For our experiments we used both a standard SVM classifier with TF-IDF features and a classifier based on the average of word embeddings within the tweet.

With regards to pre-trained LMs, we used BERT (Bidirectional Encoder Representations from Transformers) and ALBERT (A Lite BERT) \cite{lan2019albert}. These LMs have been deployed effectively in NLP tasks, leading to state-of-the-art results in most standard benchmarks \cite{wang2019glue} including Twitter \cite{basile2019semeval,roitero2020twitter}. In particular, ALBERT has been shown to provide competitive results despite being relatively light-weight compared to other LMs.

Finally, for completeness we added a naïve baseline that predicts positive instances in all cases.

\subsubsection{Training details}

We used the scikit-learn SVM model \cite{pedregosa2011scikit} as well as its TF-IDF (Term Frequency-Inverse Document Frequency) vectorizer implementations.\footnote{https://scikit-learn.org/stable/modules/generated/sklearn.feature\_extraction.text.TfidfVectorizer.html} The word embeddings generated for each tweet were drawn from vectors trained on Twitter data \cite[GloVe]{pennington2014glove}. These vectors had a dimensionality of 200, and so did the averaged embedding generated.

We performed tweet text preprocessing prior to their input to the SVM. In one series of SVM experiments all tweets underwent tokenization and lowercasing only, but in a second series all tweets also underwent tweet specific preprocessing\footnote{https://pypi.org/project/tweet-preprocessor/} (SVM+preproc henceforth). The preprocessing entailed the removal of hashtags, user mentions, reserved words (such as ``RT'' and ``FAV''), emojis, and smileys. This enabled us to see how the presence of these common tweet features affected classification performance. In both cases, the SVM used a linear kernel and default hyperparameters.

To deploy the LM classifiers we used the Simple Transformers\footnote{https://github.com/ThilinaRajapakse/simpletransformers} software library. It provides a convenient Application Programming Interface (API) to the Transformers Library, which itself provides access to BERT and ALBERT models, amongst others \cite{Wolf2019HuggingFacesTS}. The BERT and ALBERT classifiers used were ``bert-base-uncased'' and  ``albert-base-v1'',\footnote{https://huggingface.co/transformers/pretrained\_models.html} respectively.%The BERT classifier used was ``bert-base-uncased'', a model pre-trained on lower-cased English text. The ALBERT classifier used was ``albert-base-v1''\footnote{https://huggingface.co/transformers/pretrained\_models.html}.

Unlike the SVM experiments it was not necessary to tokenize tweet texts prior to their input to the BERT or ALBERT classifiers; they perform their own tokenization. Tweet texts did undergo prior lowercasing however. The classifiers were instantiated with Simple Transformers' default hyperparameters.% only, which included one training epoch.

\subsection{Results}
\label{results}

Experimental results are presented in Table \ref{table:3} (DATD) and Table \ref{table:4} (DATD+Rand).\footnote{Results for the LMs BERT and ALBERT were reported for the average of five different runs.}

\begin{table}[t!]
\centering
\begin{tabular}{ll|llll}
\hline
\textbf{Classifier} & \textbf{Features} & \textbf{Accuracy} & \textbf{Precision} & \textbf{Recall} & \textbf{F1}\\
\hline
\multirow{3}{*}{SVM} & TF-IDF & 0.633 & 0.619 & 0.693 & 0.654\\
 & Word Embs & 0.727 & 0.693 & 0.813 & 0.748\\
 & TF-IDF + Word Embs & 0.733 & 0.711 & 0.787 & 0.747\\
\cline{1-2}
\multirow{3}{*}{SVM+preproc} & TF-IDF & 0.633 & 0.616 & 0.707 & 0.658\\
 & Word Embs & 0.713 & 0.695 & 0.760 & 0.726\\
 & TF-IDF + Word Embs & 0.727 & 0.698 & 0.800 & 0.745\\
\cline{1-2}
BERT & LM & \textbf{0.749} & \textbf{0.713} & \textbf{0.856} & \textbf{0.774}\\
ALBERT & LM & 0.675 & 0.651 & 0.779 & 0.705\\
\hline
Naïve baseline & - & 0.500 & 0.500 & 1.000 & 0.670\\
\hline
\end{tabular}
\caption{Results of the first experimental setup (DATD).}
\label{table:3}
\end{table}

In the first setting, BERT achieves the best overall results, which is not unexpected. More importantly, the overall accuracy (i.e. 0.749) is close to the pairwise IAA (i.e. 0.800), which suggests that BERT is able to follow the guidelines provided in Section \ref{annotation} to a reasonable extent. As for the linear models, the Twitter-specific preprocessing for the SVM does not lead to any improvements. In the following analysis section, we aim at shedding light on the types of error made by these models. 

In the second setting where random tweets are added as negative instances in the training sets (Table \ref{table:4}), SVM with word embeddings features perform similarly to BERT, being in fact slightly better overall. This result is perhaps surprising, but may be due to the relative robustness of SVMs with respect to unbalanced training sets, which seem to have a greater effect on the LMs. Another explanation may be that by concatenating TF-IDF features and word embeddings the classifier is effectively leveraging both global and local dependencies, which have been shown to be crucial in tweet classification tasks such as emoji prediction \cite{barbieri2018semeval} and stance detection \cite{mohammad2016semeval}.

More generally, the results in this setting are not hugely different from the first setting's. This is encouraging, as it suggests that supervised models can also perform in a more realistic setting where the negative instances are more prevalent than the positive ones.

\begin{table}[t!]
\centering
\begin{tabular}{ll|llll}
\hline
\textbf{Classifier} & \textbf{Features} & \textbf{Accuracy} & \textbf{Precision} & \textbf{Recall} & \textbf{F1}\\
\hline
\multirow{3}{*}{SVM} & TF-IDF & 0.673 & 0.681 & 0.653 & 0.667\\
 & Word Embs & 0.740 & 0.737 & 0.747 & 0.742\\
 & TF-IDF + Word Embs & \textbf{0.747} & \textbf{0.740} & 0.760 & \textbf{0.750}\\
\cline{1-2}
\multirow{3}{*}{SVM+preproc} & TF-IDF & 0.660 & 0.658 & 0.667 & 0.662\\
 & Word Embs & 0.720 & 0.704 & 0.760 & 0.731\\
 & TF-IDF + Word Embs & 0.740 & 0.725 & 0.773 & 0.748\\
\cline{1-2}
BERT & LM & 0.693 & 0.656 & 0.851 & 0.737\\
ALBERT & LM & 0.648 & 0.609 & \textbf{0.880} & 0.715\\
\hline
Naïve baseline & - & 0.500 & 0.500 & 1.000 & 0.670\\
\hline
\end{tabular}
\caption{Results of the second experimental setup (DATD+Rand).}
\label{table:4}
\end{table}

\subsection{Analysis}
\label{analysis}

Perhaps the main highlights of our experiments are the results obtained by BERT and the concatenation of TF-IDF features and word embeddings in SVMs. BERT performs remarkably well despite not being trained on Twitter data. This could suggest that, although slang, jargon, misspellings, and emoji are typical in microposts, users suffering from mental illness are more articulate in their online writing than their mentally healthy counterparts. Thus their writing style is more likely to be picked up by an LM with restricted vocabularies.\footnote{While this is overcome within BERT by the use of WordPiece \cite{wu2016google}, the quality of its internal representations degrade as the number of OOV words it has to deal with increases.} Another surprising set of results concerns ALBERT, which was trained on the same corpus as BERT. %two corpora totalling over 800M words \cite{lan2019albert}. 
We could expect that given the modest size of this dataset, an overparameterized model like BERT could fall short when compared to lighter versions, but this does not seem to be the case. In any case, ALBERT achieves the highest recall score on the DATD+Rand dataset.

Let us now highlight a selection of tweets in the DATD dataset and the performance of the classifier configurations against them. For example, the tweet \emph{``got a yellow phone case hoping it will cure my depression''} was labelled \textbf{1}, a label only predicted by two of the eight configurations, namely BERT and SVM with TF-IDF + Word Embs features. This tweet is a good example of the overarching complexity of the problem, the presence of prosaic terms like ``phone case'', or positive words like ``hope'' or ``cure'' may have confused the simpler word-based models.

Another illustrative example is \emph{``you know that i'm the best, is that why you depressed?''}, which was labelled \textbf{0} and was misclassified by all configurations. We hypothesize that this may be due to having three instances of two distinct pronouns (``I'' and ``you''), which are likely used often by depressed or anxious tweeters, although it was not the case in this particular example.

There are other interesting examples, including cases where only the LMs (BERT and ALBERT) made correct predictions. For example, \emph{``Got my first call center job and my anxiety is through the roof''}, which was labelled \textbf{1}, and \emph{``Hot shot screaming gives me anxiety watching these game lol mans be stressed but these games good as hell learning a lot.''}, which was labelled \textbf{0}. Conversely, there are also cases where only SVMs made correct predictions. For example \emph{``big ass spider in my room and it disappeared so I'll just have anxiety for the rest of the night I'm in here''}, which was labelled \textbf{0}. While it is not clear whether there is a systematic pattern to draw conclusions from, it does seem that when only the LMs succeed there is some degree of world or semantic understanding required to capture the condition of the tweeter (for example, the fact that you are probably not actually anxious by watching a game).

\section{Conclusion and Future Work}

In this paper, we have presented an experimental evaluation for detecting depression and anxiety in social media. We have developed a dataset, DATD, for predicting depression and anxiety in Twitter. Using this dataset we have run a comparative analysis of pre-trained LMs and traditional linear models. Not surprisingly, LMs performed relatively well on this task with a balanced set, but they do not outperform lighter-weight methods when the training data is unbalanced. Given the relatively small size of the dataset, we have also performed a qualitative analysis to identify areas for improvement. 

%Since these automatic models are intended to be used by medical experts, as future work it would be interesting to develop a further collaboration with such experts.
Since these automatic models are intended for use by medical experts future work could involve collaboration with them. Moreover, classifier performance must be measured and certified versus large, heterogeneous datasets before adoption in healthcare is likely to be considered \cite{kelly2019key}. Collaboration may serve mental health experts to better understand social media at a large-scale, and to develop better guidelines and treatments.

Finally, we are also planning to extend this work to develop a dataset with finer grained distinctions, similar to \newcite{bathina2020depressed} for Cognitive Distortion Schemas.

%\section*{Acknowledgements}

%The acknowledgements should go immediately before the references.  Do
%not number the acknowledgements section. Do not include this section
%when submitting your paper for review.

\end{spacing}

% include your own bib file like this:
\bibliographystyle{coling}
\bibliography{coling2020}

\begin{thebibliography}{}

\bibitem[\protect\citename{Baclic \bgroup et al.\egroup
  }2020]{baclic2020challenges}
Oliver Baclic, Matthew Tunis, Kelsey Young, Coraline Doan, Howard Swerdfeger,
  and Justin Schonfeld.
\newblock 2020.
\newblock Challenges and opportunities for public health made possible by
  advances in natural language processing.
\newblock {\em Canada Communicable Disease Report}, 46(6):161--168.

\bibitem[\protect\citename{Barbieri \bgroup et al.\egroup
  }2018]{barbieri2018semeval}
Francesco Barbieri, Jose Camacho-Collados, Francesco Ronzano, Luis~Espinosa
  Anke, Miguel Ballesteros, Valerio Basile, Viviana Patti, and Horacio Saggion.
\newblock 2018.
\newblock Semeval 2018 task 2: Multilingual emoji prediction.
\newblock In {\em Proceedings of The 12th International Workshop on Semantic
  Evaluation}, pages 24--33.

\bibitem[\protect\citename{Basile \bgroup et al.\egroup
  }2019]{basile2019semeval}
Valerio Basile, Cristina Bosco, Elisabetta Fersini, Nozza Debora, Viviana
  Patti, Francisco Manuel~Rangel Pardo, Paolo Rosso, Manuela Sanguinetti,
  et~al.
\newblock 2019.
\newblock Semeval-2019 task 5: Multilingual detection of hate speech against
  immigrants and women in twitter.
\newblock In {\em 13th International Workshop on Semantic Evaluation}, pages
  54--63. Association for Computational Linguistics.

\bibitem[\protect\citename{Bathina \bgroup et al.\egroup
  }2020]{bathina2020depressed}
Krishna~C. Bathina, Marijn~ten Thij, Lorenzo Lorenzo-Luaces, Lauren~A Rutter,
  and Johan Bollen.
\newblock 2020.
\newblock Depressed individuals express more distorted thinking on social
  media.
\newblock {\em arXiv preprint arXiv:2002.02800}.

\bibitem[\protect\citename{Bolton}2008]{BoltonDerek2008Wimd}
Derek Bolton.
\newblock 2008.
\newblock {\em What is mental disorder? : an essay in philosophy, science, and
  values}.
\newblock International perspectives in philosophy and psychiatry. Oxford
  University Press, Oxford ; New York.

\bibitem[\protect\citename{{Centers for Disease Control and
  Prevention}}2015]{centers2015suicide}
{Centers for Disease Control and Prevention}.
\newblock 2015.
\newblock Suicide: Facts at a glance [fact sheet].

\bibitem[\protect\citename{{Dennis C Miller}}2016]{miller2016}
{Dennis C Miller}.
\newblock 2016.
\newblock Mental health awareness month: Take the first step towards a mentally
  healthy workplace.

\bibitem[\protect\citename{Devlin \bgroup et al.\egroup }2019]{BERT}
Jacob Devlin, Ming-Wei Chang, Kenton Lee, and Kristina Toutanova.
\newblock 2019.
\newblock Bert: Pre-training of deep bidirectional transformers for language
  understanding.
\newblock In {\em Proceedings of the 2019 Conference of the North American
  Chapter of the Association for Computational Linguistics: Human Language
  Technologies, Volume 1 (Long and Short Papers)}, pages 4171--4186.

\bibitem[\protect\citename{Guntuku \bgroup et al.\egroup
  }2017]{guntuku2017detecting}
Sharath~Chandra Guntuku, David~B Yaden, Margaret~L Kern, Lyle~H Ungar, and
  Johannes~C Eichstaedt.
\newblock 2017.
\newblock Detecting depression and mental illness on social media: an
  integrative review.
\newblock {\em Current Opinion in Behavioral Sciences}, 18:43--49.

\bibitem[\protect\citename{Han \bgroup et al.\egroup }2020]{han2020application}
Kai-Xu Han, Wei Chien, Chien-Ching Chiu, and Yu-Ting Cheng.
\newblock 2020.
\newblock Application of support vector machine (svm) in the sentiment analysis
  of twitter dataset.
\newblock {\em Applied Sciences}, 10(3):1125.

\bibitem[\protect\citename{Harris \bgroup et al.\egroup
  }2014]{harris2014health}
Jenine~K Harris, Raed Mansour, Bechara Choucair, Joe Olson, Cory Nissen, and
  Jay Bhatt.
\newblock 2014.
\newblock Health department use of social media to identify foodborne
  illness—chicago, illinois, 2013--2014.
\newblock {\em MMWR. Morbidity and mortality weekly report}, 63(32):681.

\bibitem[\protect\citename{Holzinger}2016]{holzinger2016interactive}
Andreas Holzinger.
\newblock 2016.
\newblock Interactive machine learning for health informatics: when do we need
  the human-in-the-loop?
\newblock {\em Brain Informatics}, 3(2):119--131.

\bibitem[\protect\citename{Kelly \bgroup et al.\egroup }2019]{kelly2019key}
Christopher~J Kelly, Alan Karthikesalingam, Mustafa Suleyman, Greg Corrado, and
  Dominic King.
\newblock 2019.
\newblock Key challenges for delivering clinical impact with artificial
  intelligence.
\newblock {\em BMC medicine}, 17(1):195.

\bibitem[\protect\citename{Lan \bgroup et al.\egroup }2019]{lan2019albert}
Zhenzhong Lan, Mingda Chen, Sebastian Goodman, Kevin Gimpel, Piyush Sharma, and
  Radu Soricut.
\newblock 2019.
\newblock Albert: A lite bert for self-supervised learning of language
  representations.
\newblock {\em arXiv preprint arXiv:1909.11942}.

\bibitem[\protect\citename{Lexis \bgroup et al.\egroup
  }2011]{lexis2011prevention}
Monique~AS Lexis, Nicole~WH Jansen, Marcus~JH Huibers, Ludovic~GPM
  Van~Amelsvoort, Ate Berkouwer, Gladys Tjin~A Ton, Piet~A Van Den~Brandt, and
  IJmert Kant.
\newblock 2011.
\newblock Prevention of long-term sickness absence and major depression in
  high-risk employees: a randomised controlled trial.
\newblock {\em Occupational and Environmental Medicine}, 68(6):400--407.

\bibitem[\protect\citename{Mohammad \bgroup et al.\egroup
  }2016]{mohammad2016semeval}
Saif Mohammad, Svetlana Kiritchenko, Parinaz Sobhani, Xiaodan Zhu, and Colin
  Cherry.
\newblock 2016.
\newblock Semeval-2016 task 6: Detecting stance in tweets.
\newblock In {\em Proceedings of the 10th International Workshop on Semantic
  Evaluation (SemEval-2016)}, pages 31--41.

\bibitem[\protect\citename{Pedregosa \bgroup et al.\egroup
  }2011]{pedregosa2011scikit}
Fabian Pedregosa, Ga{\"e}l Varoquaux, Alexandre Gramfort, Vincent Michel,
  Bertrand Thirion, Olivier Grisel, Mathieu Blondel, Peter Prettenhofer, Ron
  Weiss, Vincent Dubourg, et~al.
\newblock 2011.
\newblock Scikit-learn: Machine learning in python.
\newblock {\em the Journal of machine Learning research}, 12:2825--2830.

\bibitem[\protect\citename{Pennebaker \bgroup et al.\egroup
  }2015]{Pennebaker2015TheDA}
James~W. Pennebaker, Ryan~L. Boyd, Kayla~N Jordan, and Kate Blackburn.
\newblock 2015.
\newblock The development and psychometric properties of liwc2015.
\newblock In {\em Psychometrics manual for text analysis program LIWC2015}.

\bibitem[\protect\citename{Pennington \bgroup et al.\egroup
  }2014]{pennington2014glove}
Jeffrey Pennington, Richard Socher, and Christopher~D Manning.
\newblock 2014.
\newblock Glo{V}e: Global vectors for word representation.
\newblock In {\em Proceedings of EMNLP}, pages 1532--1543.

\bibitem[\protect\citename{Prieto \bgroup et al.\egroup
  }2014]{prieto2014twitter}
V{\'\i}ctor~M. Prieto, Sergio Matos, Manuel Alvarez, Fidel Cacheda, and
  Jos{\'e}~Lu{\'\i}s Oliveira.
\newblock 2014.
\newblock Twitter: a good place to detect health conditions.
\newblock {\em PloS one}, 9(1):e86191.

\bibitem[\protect\citename{Roitero \bgroup et al.\egroup
  }2020]{roitero2020twitter}
Kevin Roitero, VDMSM Cristian~Bozzato, and G~Serra.
\newblock 2020.
\newblock Twitter goes to the doctor: Detecting medical tweets using machine
  learning and bert.
\newblock In {\em Proceedings of the International Workshop on Semantic
  Indexing and Information Retrieval for Health from heterogeneous content
  types and languages (SIIRH 2020)}.

\bibitem[\protect\citename{Wang \bgroup et al.\egroup }2019]{wang2019glue}
Alex Wang, Amanpreet Singh, Julian Michael, Felix Hill, Omer Levy, and Samuel
  Bowman.
\newblock 2019.
\newblock Glue: A multi-task benchmark and analysis platform for natural
  language understanding.
\newblock In {\em 7th International Conference on Learning Representations,
  ICLR 2019}.

\bibitem[\protect\citename{Wasserman \bgroup et al.\egroup
  }2012]{wasserman2012suicide}
Camilla Wasserman, Christina~W Hoven, Danuta Wasserman, Vladimir Carli, Marco
  Sarchiapone, Susana Al-Halabi, Alan Apter, Judit Balazs, Julio Bobes, Doina
  Cosman, et~al.
\newblock 2012.
\newblock Suicide prevention for youth-a mental health awareness program:
  lessons learned from the saving and empowering young lives in europe (seyle)
  intervention study.
\newblock {\em BMC public health}, 12(1):776.

\bibitem[\protect\citename{Wolf \bgroup et al.\egroup
  }2019]{Wolf2019HuggingFacesTS}
Thomas Wolf, Lysandre Debut, Victor Sanh, Julien Chaumond, Clement Delangue,
  Anthony Moi, Pierric Cistac, Tim Rault, R'emi Louf, Morgan Funtowicz, and
  Jamie Brew.
\newblock 2019.
\newblock Huggingface's transformers: State-of-the-art natural language
  processing.
\newblock {\em ArXiv}, abs/1910.03771.

\bibitem[\protect\citename{{World Health Organization}}2016]{world2016world}
{World Health Organization}.
\newblock 2016.
\newblock {\em World health statistics 2016: monitoring health for the SDGs
  sustainable development goals}.
\newblock World Health Organization.

\bibitem[\protect\citename{Wu \bgroup et al.\egroup }2016]{wu2016google}
Yonghui Wu, Mike Schuster, Zhifeng Chen, Quoc~V Le, Mohammad Norouzi, Wolfgang
  Macherey, Maxim Krikun, Yuan Cao, Qin Gao, Klaus Macherey, et~al.
\newblock 2016.
\newblock Google's neural machine translation system: Bridging the gap between
  human and machine translation.
\newblock {\em arXiv preprint arXiv:1609.08144}.

\bibitem[\protect\citename{Xie \bgroup et al.\egroup
  }2017]{xie2017reformulation}
Zheng Xie, Chaitanya Gadepalli, and Barry~MG Cheetham.
\newblock 2017.
\newblock Reformulation and generalisation of the cohen and fleiss kappas.
\newblock {\em LIFE: International Journal of Health and Life-Sciences}, 3(3).

\bibitem[\protect\citename{Yates \bgroup et al.\egroup
  }2017]{yates2017depression}
Andrew Yates, Arman Cohan, and Nazli Goharian.
\newblock 2017.
\newblock Depression and self-harm risk assessment in online forums.
\newblock In {\em Proceedings of the 2017 Conference on Empirical Methods in
  Natural Language Processing}, pages 2968--2978.

\bibitem[\protect\citename{Zapf \bgroup et al.\egroup }2016]{zapf2016measuring}
Antonia Zapf, Stefanie Castell, Lars Morawietz, and Andr{\'e} Karch.
\newblock 2016.
\newblock Measuring inter-rater reliability for nominal data--which
  coefficients and confidence intervals are appropriate?
\newblock {\em BMC medical research methodology}, 16(1):93.

\end{thebibliography}

%\begin{thebibliography}{}

%\end{thebibliography}

\end{document}